\documentclass[runningheads]{llncs}
\usepackage{graphicx}
\usepackage{bm}
\usepackage{bbm}
\usepackage{amssymb}
\usepackage{amsmath}
\usepackage{subfigure}
\usepackage{multirow}
\usepackage{booktabs}
\usepackage{subfigure}
\begin{document}
\title{Time-Guided High-Order Attention Model of Longitudinal Heterogeneous Healthcare Data}
\titlerunning{TGHOA Model of Longitudinal Heterogeneous Healthcare Data}

\author{Yi Huang\inst{1,2} \and
Xiaoshan Yang\inst{1,2} \and
Changsheng Xu\inst{1,2}}
%\author{Anonymous Author(s)}
%
\authorrunning{Y. Huang et al.}
%\authorrunning{Anonymous Author(s)}
\institute{Institute of Automation, Chinese Academy of Sciences, Beijing, China\and
University of Chinese Academy of Sciences, Beijing, China\\
\email{\{yi.huang,xiaoshan.yang,csxu\}@nlpr.ia.ac.cn}}

\maketitle              % typeset the header of the contribution
\begin{abstract}
Due to potential applications in chronic disease management and personalized healthcare, the EHRs data analysis has attracted much attentions of both researchers and practitioners. There are three main challenges in modeling longitudinal and heterogeneous EHRs data: heterogeneity, irregular temporality and interpretability. A series of deep learning methods have made remarkable progress in resolving these challenges. Nevertheless, most of existing attention models rely on capturing the 1-order temporal dependencies or 2-order multimodal relationships among feature elements. In this paper, we propose a time-guided high-order attention (TGHOA) model. The proposed method has three major advantages. (1) It can model longitudinal heterogeneous EHRs data via capturing the 3-order correlations of different modalities and the irregular temporal impact of historical events. (2) It can be used to identify the potential concerns of medical features to explain the reasoning process of healthcare model. (3) It can be easily expanded into cases with more modalities and flexibly applied in different prediction tasks. We evaluate the proposed method in two tasks of mortality prediction and disease ranking on two real world EHRs datasets. Extensive experimental results show the effectiveness of the proposed model.
%\keywords{First keyword  \and Second keyword \and Another keyword.}
\end{abstract}
%
%
%


\section{Introduction}
%先有数据，然后可以有什么应用？过去的方法有什么问题？深度学习技术兴起了，可以用到这个领域吗？有什么特殊问题要解决（3点）？为此我们提出了什么、解决了什么。总结一下优点。
With the wide use of digital devices and information systems in hospital, a large volume of Electronic Health Records (EHRs) data have been accumulated during the patients' admissions to the hospital. EHRs consist of sequential records such as diagnoses, physical test indicators and medication prescriptions.
Due to potential applications in chronic disease management and personalized healthcare, such EHRs data have attracted remarkable attentions of both researchers and practitioners. 
Deep learning based methods are widely used to model EHRs data in healthcare tasks, including disease detection ~\cite{ma2017dipole,thodoroff2016learning,jagannatha2016structured}, medical concept embedding ~\cite{choi2016multi,cai2018medical}, computational phenotyping \cite{baytas2017patient,suresh2017use,che2017rnn} and clinical event prediction \cite{choi2016doctor,xu2018raim,jagannatha2016structured}.
However, it is still challenging to improve the quality and efficiency of the healthcare/disease management by mining large-scale heterogeneous EHRs data, where the treatment records provided by senior doctors and physical examination results monitored during hospital staying always have different formats with various recording frequencies.

%enable researchers to build computer information systems for patient health analysis.
%With the rapid development of artificial intelligence and deep learning over the past few years, dramatic performance improvements have been obtained in many data analytic tasks such as computer vision and natural language processing.
%{\color{red}please add explanation of longitudinal data}.

%It is challenge but potentially beneficial to develop more accurate model for healthcare analysis by leveraging knowledge learned from such two types of longitudinal heterogeneous EHRs data.

There are three challenges in modeling the vast amount of longitudinal heterogeneous EHRs data: 
%There are two challenges to solve the above problem.
%First, the recorded data in EHRs sequences are provided in an unstructured form.
(1) \textit{Heterogeneity}:
EHRs data are collected from multiple devices and monitors.
Multiple data streams are recorded
for different destinations in different forms.
For example,
during a patient's hospital stay, primary diagnostic codes are recorded by doctors for developing treatment plan, while some physical examination results are recorded by medical instruments for monitoring and evaluating the patient's conditions.
(2) \textit{Irregular temporality}:
On the one hand,
the  diagnostic codes and physical indicators are always sampled at different frequencies (e.g., ECG sampled dozens per second and vital signs sampled minutely).
Moreover, the varying length of hospital staying also leads to the different length of the record sequence in different hospital visits.
On the other hand, for a patient with multiple hospital visits, the time interval between two consecutive visits can vary from days to months.
%It requires a robust method to integrate such heterogeneous data series.
	%The first consideration is the heterogeneity of EHRs data. For example, at each admission for a patient, discrete diagnostic codes are made once by doctors, while the lab indicators, which are tested by medical devices with different frequencies during hospitalization, are temporally ordered. Another notable characteristic of EHRs event sequence is the irregularity of time interval between two consecutive hospital visits, which can vary from days to months. Noting that all these data are recorded during a patient's long-term treatment, it requires a robust method to leverage these continuously and inconsistently available multi-source data.
	%as the time intervals at which the health events occurred are irregular and can vary from days to months. The inputs of traditional sequence tasks such as sentence translation usually have a fixed sequential interval. However,
%For example, most of diseases have different progression over the same time. On the other hand, a disease obviously has different progression after different passing time.
	%Meanwhile, all these kinds of medical factors would affect each other. For example, the diagnostic information describes a patient's health condition and determines the following treatment plan and what laboratories should the patient be examined regularly, while the dynamic data (i.e., lab indicators) reflect more about the effect of treatment on the patient.
	%措辞修改
(3) \textit{Interpretability}:
It is important to improve the interpretability of the healthcare analysis model in addition to the prediction performance on EHRs sequence data.
%Though deep learning based sequential models have obtained successful predictive results
%on the EHRs data
%{\color{red}~\cite{XXX}},
%they are always applied as a black
%box in many healthcare tasks.
%it is also important to keep a balance between accuracy and interpretability.
To help doctors and patients with a lot of complex EHRs data,
a natural requirement is to identify the supporting evidences for the conclusions.
%An interpretable model needs to identify the distinguished medical features which contribute most to the final prediction result.

Over the past few years, a series of deep learning methods have made remarkable progress in resolving these challenges.
Existing models often make efforts on improving the prediction performance by capturing the sequential manner of the EHRs data \cite{choi2016using,baytas2017patient}, or representing the recorded medical concepts\cite{miotto2016deep,choi2016multi}.
%change these refers to methods without attention
%These methods are always applied as a black box in many healthcare tasks.
In order to getting interpretable results,
%some attention-based models have been applied on EHRs data via calculating a special form of input for particular tasks.
attention-based models geared towards a specific form of input for a
particular task.
~\cite{choi2017gram} learns medical concepts with external knowledge.
~\cite{choi2016retain,xu2018raim} learn to selectively attend on different medical features.
%change these refers to methods without attention
%the opportunity to resolve both challenges is likely to derive from
%While interpretable, existing attention mechanisms~\cite{choi2016retain,xu2018raim} are often geared
%towards a specific form of input for a particular task.
Most of these models rely on aggregated features via
capturing the 1-order temporal dependencies or 2-order multimodal relationships among feature elements of the EHRs data.

In fact, when evaluating patients' health condition,
a doctor would comprehensively review both the past medical records and the current reports
to find correlation factors,
then focus on specific medical features, and make their decisions finally.
This kind of reasoning process
simultaneously explores the correlations of multiple
data sources, such as medical diagnoses, lab indicators and
the history medical events.
%As a consequence, the accuracy results of existing attention models are sub-optimal.
%To effectively model the heterogeneous EHRs data sequences,
Since most of existing attention models in healthcare
only consider 1-order or 2-order relationships,
the opportunity is likely to derive from
%we propose a attention mechanism
learning high-order correlations (3-order and above) among feature elements.
Learning these
correlations effectively directs the appropriate attention to the relevant elements in different data
modalities and at different time steps that are required to jointly solve the prediction task.
In this paper,
we propose a Time-Guided High-Order Attention (TGHOA)  model for analyzing the
heterogeneous and irregular temporal longitudinal EHRs data.
The proposed TGHOA jointly models the
correlations of different types of longitudinal medical records
and the irregular temporal impact of historical events.
Specifically,
%for each visit of the patient,
we compute the one-hot medical diagnose feature by
embedding scheme.
The uniform representations
of physical indicators with different recording frequencies and lengths
are computed by convolution kernels.
The diagnose features, physical indicator features and historical event features
are comprehensively used to
compute a relationship matrix
which is further transformed to
attention scores.
Considering that a larger time interval between the previous visit
and the current visit leads to less impact of the historical event feature,
the time gap is
used as an important factor to guide the attention computing.
Finally, the attended features
are combined together to predict patient's health.
%
%The proposed TGHOA is a general
%attention mechanism
%that can be trained in an end-to-end form
%in general neural networks.
Figure~\ref{fig:all} shows the framework of the proposed method.
%
%with three units:
%1) temporal dependency of each type of current data,
%2) correlations between different types of current data,
%and 3) high-order correlations between historical treatment records and current heterogeneous data.
%	1) a representation learning module for building heterogeneous data;
%	2) a High-Order Attention module for capturing both temporal dependencies within EHRs sequential data and inter-correlations between EHRs data;
%	and 3) a LSTM module for sequential encoding.
%	

	%Compared with previous healthcare models of  EHRs data, our proposed model not only try to learn what medical factors to pay attention to, but also can discover temporal dependency level of each source data on historical treatment records, and show how the current heterogeneous medical features correlate with each other.
	
To summarize, the main contributions of this paper are as follows:
\begin{itemize}		
\item
The proposed method can model longitudinal heterogeneous EHRs data via an efficient 3-order attention mechanism, to simultaneously capture the correlations of different modalities
and the irregular temporal impact of historical events.%with considering of irregular temporality.
		
\item
The proposed high-order attention module can be used to identify the potential concerns of medical features to explain the reasoning process of healthcare model.
		% Simultaneously, it could discover temporal dependency of each source data on historical treatment records, and show how multi-source medical factors correlate with each other. These functions are helpful to explain the reasoning process of healthcare model.
		
\item
Due to the efficient computation formula of the proposed higher-order attention mechanism,
it can be easily expanded into cases with more modalities
and flexibly applied in different
prediction tasks.
In our work, we evaluate the proposed method in two tasks of mortality prediction and disease ranking
on two real world EHRs datasets.
\end{itemize}

\section{Related Work}
Traditional health analysis system often depends on labor intensive efforts, such as expert-defined phenotyping \cite{richesson2016clinical,pathak2013electronic} and manual feature engineering \cite{xu2012feature}. We briefly review the three kinds of deep learning based methods mostly related to our work.
%temporality

\noindent\textbf{Deep Learning on Longitudinal EHRs data.}
~\cite{lipton2015learning} shows that RNN models, which can capture the dynamic relationships in sequential data, perform pretty good in large historical data of EHRs.
%	\cite{choi2016retain} used two RNNs to model visit-level and variable-level attention mechanisms and selectively attend to medical variable within past visits.
In addition, \cite{che2017rnn} found that the irregularity of longitudinal EHRs data would affect model performance and used Dynamic Time Warping (DTW) to match irregular temporal patterns in data sequences.
\cite{baytas2017patient} proposed a novel LSTM architecture, which performs a subspace decomposition module and a time-decaying memory module followed by the standard gated architecture of LSTM, to handle time irregularities in sequences.
These methods do not consider hidden inter-correlation between different medical variables in heterogeneous EHRs data and lack of interpretability.

%Heterogeneity
\noindent\textbf{Deep Learning on Heterogeneous EHRs data.}
%by exploring correlation patterns among different types of medical sequences.
~\cite{jin2018treatment} designed a heterogeneous LSTM structure to explore multiple inter-correlations of different medical sequences with different lengths and record frequencies.
~\cite{xu2018raim} proposed an efficient multi-channel attention model of multimodal EHRs time series.
However, these models only focus on an instance encounter
%to analyze the heterogeneity of EHRs data
and do not consider the longitudinally historical records of patients.
%Such a limitation that the longitudinal EHRs records of a patient are treated as independent visits affects accuracy and utility of the healthcare analysis.

\noindent\textbf{Attention-based Interpretable Deep Methods.}
%Attention mechanism is a useful method which launch interoperability into deep model for video understanding and machine translation. Attention has also been employed to models for EHRs Data.
RETAIN \cite{choi2016retain} used two RNNs to model visit-level and variable-level attention mechanisms. Thus it could determine which visit and which medical variable it should pay attention while doing predicting.
GRAM \cite{choi2017gram} used a graph-based attention model in two sequential diagnoses prediction tasks and one heart failure (HF) prediction task. This method could learn robust representations of medical code via a knowledge graph which describes medical ontology relationships.
%~\cite{sha2017interpretable} proposed another hierarchical attention model to incorporate external information into the EHRs models.
RAIM \cite{xu2018raim} proposed a recurrent attentive and intensive model for analyzing the multimodal EHR time series. RAIM uses an efficient multi-channel attention on continuous monitored data, which is guided by discrete clinical data.
Different from these works, we design a high-order attention module to jointly handle the
irregular temporality and heterogeneity of the EHRs data.

\section{Methods}\label{sec:methods}
\begin{figure*}[t]
	\centering
	\includegraphics[width=12cm]{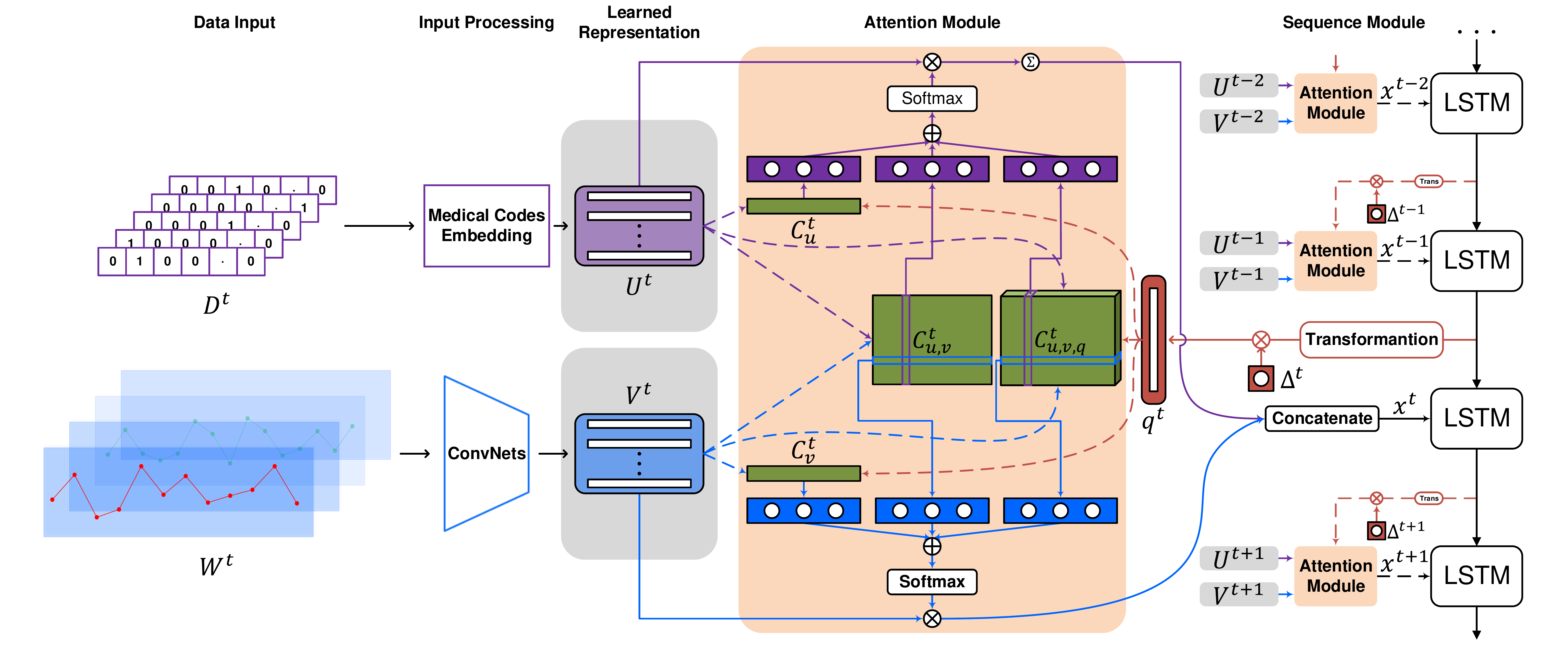}
	\caption{An overview of the proposed model. }
	\label{fig:all}
\end{figure*}

In this section, we first define the notations describing
the original EHRs events sequence, followed by
representation methods of two types of heterogeneous sequential data.
Then we describe the details of the proposed time-guided high-order attention module.
Finally, we introduce the decision-making process based on the attended features.
%In this paper,
%% as our method is easy to be extended to consider more data sources,
%we mainly consider the high-order correlation within two types of heterogeneous data in longitudinal EHRs.
Figure~\ref{fig:all} shows an overview of our method.
%As shown in Figure \ref{fig:all}, our proposed model fuses two types of sequential data withe time-guided memory query via a High-Order Attention module and feeds weighted input features into LSTM for making predictions.

\subsection{Notations}\label{sec:notations}
To reduce clutter, we will introduce our method for a single patient.
We define a patient's $t$-th visit to hospital as one EHRs event $\mathcal{E}^t$,
and  multiple visits are denoted as a EHRs event sequence $\mathcal{P} = \{\mathcal{E}^1, \mathcal{E}^2, \ldots , \mathcal{E}^t, \ldots , \mathcal{E}^T\}$ where $T$ is the number of this patient's all visits.
Each visit $\mathcal{E}^t = \{D^t, W^t\}$
where $D^t$ is an integrated set of discrete diagnoses data indicating
what diseases are the patient suffering from.
$W^t$ is a set of lab results, such as \textit{saturation of pulse $O_2$} and \textit{arterial blood pressure}.
$\bm{y}$ is patient's groundtruth health evaluation after the $T$ visits.
In the experiment,
$\bm{y}$ is death rate in the mortality estimation task
and the grade of diseases in the disease ranking task.
	
Patients would be diagnosed with different diseases,
so the number of elements in $D^t$ is varying in different visits.
%We define this difference as visit-level variation.
We denote $D^t = \{\bm{d}_1^t, \bm{d}_2^t, \ldots , \bm{d}_{n_u}^t\}$,
where $\bm{d}_n^t \in \mathbb{R}^{|\mathcal{D}|}$  is a one-hot representing of patient's $n$-th disease in $t$-th visit. The $\mathcal{D}$ denotes the medical code set.
The $|\mathcal{D}|$ is the number of unique medical codes of diseases.
The $n_u$ is the number of diseases that the patient is suffering from.
The lab indicator set $W^t = \{\bm{w}_1^t, \bm{w}_2^t, \ldots, \bm{w}_{n_v}^t\}$, where $\bm{w}_i^t$ denotes $i$-th
lab indicator at the $t$-th visit of the patient.
%These lab indicators,
%such as electrocardiogram (ECG),
%can be regarded as waveforms which are much more complex than one-hot medical codes.
%Because they are measured at different frequencies
%and the lengths of them are different in each patient's visit.
%We define this difference as variable-level variation.
%What's more,
%the length of the same lab indicator is also varying in different visits
%due to the varying length of the patient's hospital staying.
\subsection{EHRs Data Representation}\label{sec:inputs}
In this section, we introduce how to to represent two types of heterogeneous data respectively.
The expressive data representations
are very important for capturing their correlation patterns.

\subsubsection{Diagnose Embedding}
Given a medical code representation  $\bm{d}_n^t \in \mathbb{R}^{|\mathcal{D}|}$, we can obtain its embedding representation $\bm{u}_n^t \in \mathbb{R}^{d_u}$ as follows:
\begin{equation}
\bm{u}_n^t = {\Theta}_{e}\bm{d}_n^t
\end{equation}
where ${\Theta}_e \in \mathbb{R}^{d_u \times |\mathcal{D}|}$ is a learned embedding matrix
and $d_u$ is the dimension of the embedding vector.
Thus the diagnostic information $D^t$ is transformed to
$U^t = \{\bm{u}_1^t, \bm{u}_2^t, \ldots , \bm{u}_{n_u}^t\}$.

\subsubsection{Lab Indicator Feature Extracting}

As mentioned in Section \ref{sec:notations} that lengths of lab indicator waveforms are different
within a single visit.
Besides, a lab indicator has different length in multiple visits.
To uniformly represent these indicator waveforms,
we design a one-dimension convolutional neural network to extract
the fixed length features:
\begin{eqnarray}
	\bm{v}_i^t = f_i(\bm{w}_i^t)
\end{eqnarray}
where $f_i(\cdot)$ is a two-layer convolutional neural network. The first convolutional block consists of a convolution layer, a max-pooling layer and a activation function ReLu. The second convolutional block consists of a convolution layer and a max-overtime pooling layer \cite{kim2014convolutional}, which is applied to naturally deals with variable waveform lengths.
So we could get the feature representation $\bm{v}_i^t \in \mathbb{R}^{d_v}$ of the lab indicator $\bm{w}_i^t$ with a fixed-length.
For different lab indicators, we initialize different network parameters of $f_i(\cdot)$ to compute their features respectively.
And the network parameters are shared among different visits for the same lab indicator.
Then the final feature of $n_v$ lab indicators in $t$-th visit are represented as
$V^t = \{\bm{v}_1^t, \bm{v}_2^t , \ldots, \bm{v}_{n_v}^t\}$.

\subsection{Time-Guided High-Order Attention}
%This section introduces how to construct an attention module to capture the high-order correlations among two types of EHRs sequential data.
In the following parts,
we will refer to the iteration of LSTM with a single step using notations as follows:
		\begin{equation}
			\bm{h}^t, \bm{c}^t = \mathtt{LSTM}(\bm{x}^t, \bm{h}^{t-1}, \bm{c}^{t-1})	\label{eq:Model}
		\end{equation}
		where $\bm{h}^t \in \mathbb{R}^{d}$ is the LSTM hidden state vecotor, $\bm{c}^t \in \mathbb{R}^{d}$ is the LSTM memory cell vecotor and $\bm{x}^t$ is the LSTM input vector which contains the information of $U^t$ and $V^t$. Here we use $d$  to denote the dimensionality of hidden vectors.
		
%Besides making the RNN models more interpretable, attention modules are a crucial component to solve many tasks with multimodal data now. \cite{schwartz2017high,anderson2018bottom} show that attention modules can learn a more succinct representation for multimodal data with high-order correlations.
Subsequently we consider the attention mechanism as an importance model with each part
computing ``importance'' of medical variable from each types of data.
We use $\lambda_{u,q}$ and $\lambda_{v,q}$ to denote the intra-sequence temporality of two types of sequential data.
$\lambda_{u,v}$ expresses inter-sequence  correlation between two data sequences.
$\lambda_{u,v,q}$ captures third-order correlation among
two types of sequential data and the history event feature.
We compute the importance scores $\bm{\alpha}_u$ and $\bm{\alpha}_v$
of the medical diagnose representations and the lab indicator features
by combination of intra-sequence irregular temporality unit, inter-sequence correlation unit, third-order correlation unit:
		\begin{equation}
			\begin{split}
				\bm{\alpha}_u^t(i_u) &= \sigma\big(\eta_1\lambda_{u,q}^t(i_u) + \eta_2\lambda_{u,v}^t(i_u) + \eta_3\lambda_{u,v,q}^t(i_u)+\eta_4\big)	\\
				\bm{\alpha}_v^t(i_v) &= \sigma\big(\varepsilon_1\lambda_{v,q}^t(i_v) + \varepsilon_2\lambda_{u,v}^t(i_v) + \varepsilon_3\lambda_{u,v,q}^t(i_v)+\varepsilon_4\big)	\label{eq:final weight}
			\end{split}
		\end{equation}
here, $\eta_i$ and $\varepsilon_i$ are learned parameters and $\sigma(\cdot)$ refers to the Softmax operation over $i_u \in \{1, \ldots, n_u\}$ and $i_v \in \{1, \ldots, n_v\}$ respectively. Such a linear combination of units provides extra flexibility for the model, since it can learn the reliability of the unit from the data.
	
\subsubsection{Intra-sequence Irregular Temporality}\label{subsubsec:intra_seq}
The intra-sequence irregular temporality unit is designed to
calculate the importance of medical factors from intra-sequence data based on
the historical event feature.
%As the attention to current medical factors is related to historical records, the inputs to the intra-sequence irregular temporality attention module are historical medical visit memory and the data representation of a specific data sequence (diagnostic codes or lab indicators).
We first define attention query $\bm{q}^t$ as
the nonlinearly transformed feature of the
previous memory $\bm{c}^{t-1}$ by a one-layer neural network.
What's more, considering that the reference value of historical records would change over time, we use a decaying function $g(\Delta^t) = 1 / \log(\mathrm{e} + \Delta^t)$ \cite{pham2016deepcare,baytas2017patient} as time guidance to adjust impact of historical memory.
$\Delta^t$ is an irregular time interval between two neighborhood visits.
So the memory query is obtained via:
			\begin{equation}\label{eq:qt}
				\bm{q}^t = g(\Delta^t)\tanh(\Theta_d\bm{c}^{t-1} + b_d)
			\end{equation}
		where $\Theta_d$ and $b_d$ are learned parameters.
			
The intra-sequence irregular temporality attention weights are formally formulated as:
			\begin{equation}
				\begin{split}
					\lambda_{u,q}^t(i_u) &= \tanh\big((\Theta_{u_1}\bm{u}_{i_u}^t)^\mathrm{T}\Theta_{u,q}\bm{q}^t\big)\\
					\lambda_{v,q}^t(i_v) &= \tanh\big((\Theta_{v_1}\bm{v}_{i_v}^t)^\mathrm{T}\Theta_{v,q}\bm{q}^t\big)
				\end{split}
			\end{equation}
			where $\Theta_{u_1}\in \mathbb{R}^{d\times d_u}$, $\Theta_{v_1}\in\mathbb{R}^{d\times d_v}$, $\Theta_{u,q}$ and $\Theta_{v,q} \in \mathbb{R}^{d\times d}$ are trainable parameters.
			%These potentials are used to compute the attention probabilities as defined in Eq. (\ref{eq:final weight}).
		
		\subsubsection{Inter-sequence Correlation}
			Besides the mentioned temporal dependencies of each data sequence, we now introduce a inter-sequence correlation unit, which is able to learn the correlation between the representations of two data sequences.
We use a relationship matrix $C_{u,v}$ between data sequences $U^t$ and $V^t$, where each entry is calculated as follows:
			\begin{equation}
				C_{u,v}^t(i_u,i_v) = (\Theta_{u_2}\bm{u}_{i_u}^t)^\mathrm{T}\Theta_{v_2}\bm{v}_{i_v}^t.
			\end{equation}
The $\Theta_{u_2}\in \mathbb{R}^{d\times d_u}$ and $\Theta_{v_2}\in\mathbb{R}^{d\times d_v}$ are trainable parameters. $C_{u,v}^t(i_u,i_v)$ measures the correlation between
the $i_u$-th diagnostic code and the $i_v$-th lab indicator. Therefore, to retrieve the attention for a specific diagnostic code or lab indicator, we convolve the matrix along the corresponding feature dimension using a $1\times1$ dimensional kernel. Specifically,
			\begin{equation}
				\begin{split}
					\lambda_{u,v}^t(i_u) = \tanh\big(\sum_{i_v=1}^{n_v}\bm{\theta}_{v_2}(i_v)C^t_{u,v}(i_u,i_v)\big)	\\
					\lambda_{u,v}^t(i_v) = \tanh\big(\sum_{i_u=1}^{n_u}\bm{\theta}_{u_2}(i_u)C^t_{u,v}(i_u,i_v)\big)
				\end{split}
			\end{equation}
			where $\bm{\theta}_{v_2} \in \mathbb{R}^{n_v}$ and $\bm{\theta}_{u_2} \in \mathbb{R}^{n_u}$ are trainable parameters.
			%These potentials are used to compute the attention probabilities as defined in Eq. (\ref{eq:final weight}).
		
\subsubsection{Time-guided Inter-sequence Correlation}
We formulate the high-order correlation between historical records and all data sequences
as follows:
			\begin{equation}
				C_{u,v,q}^t (i_u,i_v)= (\Theta_{u_3}\bm{u}_{i_u}^t\odot \Theta_q\bm{q}^t)^\mathrm{T}\Theta_{v_3}\bm{v}_{i_v}^t
			\end{equation}
			where $\Theta_{u_3}\in \mathbb{R}^{d\times d_u}$, $\Theta_{v_3}\in\mathbb{R}^{d\times d_v}$ and $\Theta_{q}\in \mathbb{R}^{d\times d}$ are trainable parameters.
Similar to the inter-sequence correlation unit, we use the relationship matrix $C_{u,v,q}^t (i_u,i_v)$ to
compute correlated attention scores for each data sequence:
			\begin{equation}
				\begin{split}
					\lambda_{u,v,q}^t(i_u) = \tanh\big(\sum_{i_v}^{n_v}\bm{\theta}_{v_3}(i_v)C_{u,v,q}^t (i_u,i_v)\big)	\\
					\lambda_{u,v,q}^t(i_v) = \tanh\big(\sum_{i_u}^{n_u}\bm{\theta}_{u_3}(i_u)C_{u,v,q}^t (i_u,i_v)\big)
				\end{split}
			\end{equation}
where $\bm{\theta}_{v_3} \in \mathbb{R}^{n_v}$ and $\bm{\theta}_{u_3} \in \mathbb{R}^{n_u}$ are trainable parameters.
%These potentials are used to compute the attention probabilities as defined in Eq. (\ref{eq:final weight}).

\subsection{Prediction Model}
After obtaining attention scores $\bm{\alpha}_u(i_u)$ and $\bm{\alpha}_v(i_v)$ for medical diagnoses and
lab indicators,
the attended features of different data sequences can be calculated respectively.
We obtain the final representation of medical codes via attentive mean-pooling as following:
		\begin{equation}
				\hat{\bm{u}}^t = \sum_{i_u=1}^{n_u}{\bm{\alpha}_u(i_u)\bm{u}_{i_u}^t} \label{eq:attended features1}
		\end{equation}
For all features of lab indicators,
we concatenate them with attention weights:
		\begin{equation}
			\hat{\bm{v}}^t = \bm{\alpha}_v(1)\bm{v}_{1}^t\oplus\bm{\alpha}_v(2)\bm{v}_{2}^t\oplus\ldots\oplus\bm{\alpha}_v(n_v)\bm{v}_{n_v}^t. \label{eq:attended features2}
		\end{equation}
We further concatenate the attended medical diagnose feature $\hat{\bm{u}}^t$ and lab indicator feature $\hat{\bm{v}}^t$ and get $
			\bm{x}^t = [\hat{\bm{u}}^t, \hat{\bm{v}}^t].	\label{eq:final attended feature}$
Then, we feed $\bm{x}^t$ as input into the LSTM sequence model described in Eq. (\ref{eq:Model}). After obtaining the final state $\bm{h}^T$, the estimated distribution over possible patient's health evaluation $\bm{y}$ is given by:
		\begin{equation}
			\bm{\hat{y}} = \mathrm{Softmax}(\Theta_of_o(\bm{h}^T) + b_o)
		\end{equation}
		where $f_o(\cdot)$ a fully-connected layer followed by activation function $\mathrm{ReLu}$.
The $\Theta_o$ and $b_o$ are learnable parameters of the output layer.
		
The parameters of all modules
%including diagnose embedding, lab indicator feature extracting,
%high-order attention, LSTM networks and final fully-connected classifier,
are trained end-to-end together by minimizing the following cross entropy loss:
		$
			\mathcal{L} = - \bm{y}^\mathrm{T}\log\bm{\hat{y}} + (1-\bm{y})^\mathrm{T}\log(1-\bm{\hat{y}}).
		$

\section{Experiments}

%In this work, we evaluate the performance of the proposed model on two real world datasets. First we describe the datasets and then compare the predictive performance with various baseline model. Finally, we qualitatively show the interpretability of our model.

\subsection{Data}
In our experiment, we adopt two real world EHRs datasets, namely MIMIC-III~\cite{johnson2016mimic} and PPMI~\cite{dinov2016predictive}.
For the MIMIC-III dataset,
the proposed high-order attention model
is applied to a binary classification task of predicting
whether the patient would die or survive in ICU.
For the PPMI dataset, the proposed attention model
is applied in prediction of disease ranking.

\subsubsection{MIMIC-III Dataset}
Medical Information Mart for Intensive Care III (MIMIC-III) is a publicly available multimodal EHRs dataset comprising deidentified health data associated with critical care patients in Beth Israel Deaconess Medical Center over 11 years \cite{johnson2016mimic}.
The data contains vital signs, laboratory measurements, diagnostic codes, survival data
of 46,520 patients.
In the mortality prediction task,
we only consider a subset of this dataset.
We extract data of patients who have more than two hospital visits.
In order to acquire better generalization ability,
we choose 1,629 diagnostic codes, whose total frequency of occurrence is greater than 95\% in the dataset.
For the lab indicators, we choose \textit{heart rate}, \textit{saturation of pulse $O_2$}, \textit{blood glucose} and \textit{arterial blood pressure} from the CHARTEVENTS table as primary physical examination data.
We finally get 9,171 records of 2,348 patients.
We randomly split the dataset into training and testing sets
with a ratio of 4:1.
The groundtruth mortality rate in the pre-processed dataset is about 22.7\%.

\subsubsection{PPMI Dataset}
Parkinson's Progression Markers Initiative (PPMI) is an observational clinical and longitudinal
{study comprising evaluations of people with Parkinson's disease (PD),
those people with high risk, and those who are healthy~\cite{dinov2016predictive}.
We refer to~\cite{che2017rnn} for data  pre-processing.
In our experiments, we use medication prescriptions as medical codes
and choose 318 physical examination features as lab indicators according to~\cite{van2011clinical}.
As a result, we get 13,768 records of 586 patients.
We randomly split the dataset into training and testing sets
with a ratio of 4:1.
For the groundtruth labels,
we use Hoehn and Yahr (NHY) scale scores \cite{hoehn1998parkinsonism} which describe how the motor functions of PD patients deteriorate.
%			 \begin{figure*}[ht]
%			 	\subfigure{\begin{minipage}{\linewidth}
%		\centering
%		\includegraphics[width=\linewidth]{a}
%		\caption{A Case Study of a Surviving Patient. }
%		\label{fig:a}\end{minipage}
%	}
%	\subfigure{
%		\begin{minipage}{\linewidth}
%		\centering
%		\includegraphics[width=\linewidth]{b}
%		\caption{A Case Study of a Dead Patient.}
%		\label{fig:b}\end{minipage}
%	}
%	\end{figure*}

\begin{table}[t]
	\centering
	\caption{Performance Comparison of Models on Prediction Task}
	
	\label{tab:performance results}\scalebox{0.8}{
		\begin{tabular}{lccccccc}
			\toprule
			{Model}			&				\multicolumn{3}{c}{MIMIC-III}										&	&\multicolumn{3}{c}{PPMI}							\\
			\cmidrule{2-4}\cmidrule{6-8}							
			&	Accuracy	&	AUC-PR	&	AUC-ROC		&&		Accuracy	&	AUC-PR		&		AUC-ROC				\\
			\midrule
			{LSTM}										&	0.7790		&	0.8520		&	0.8555	&&		0.8319		&	0.8669		&	0.9595	\\
			{LSTM-Att}								   &	0.7811		&	0.8710	   &   0.8766 &&      0.8319	   &   0.9180     &	0.9747		\\
			{T-LSTM} 								   &	0.7854		&	0.8643		&	0.8643	&&     	0.8230	   &   0.8998     &	0.9660		\\
			RETAIN								&	0.8047		&	0.8704		&	0.8772	&&     	0.8584	   &   0.9213     &	0.9755	\\
			\midrule
			{LSTM+TGA}								  &	   0.7961		&	0.8769		&	0.8829	&&  	0.8407	   &   0.9185     &	0.9764		\\
			{LSTM+CoA}	   		 				      &	   0.7876		&	0.8602		&	0.8664	&&	  0.8496	   &   0.9312     &	0.9808		\\
			{TGCoA}							     &	  0.8062	  &	   0.8878	   &   0.8867	&&	 0.8673	   &   0.9408     &	0.9837		\\
			{TGHOA}										   &	\textbf{0.8155}	    &	  \textbf{0.9091}	  &   \textbf{0.9071}	&& 	\textbf{0.8938}	   &   \textbf{0.9581}     &	\textbf{0.9883}		\\
			\bottomrule
	\end{tabular}}
\end{table}

\subsection{Implementation}	
All the model parameters introduced in Section~\ref{sec:methods} are randomly initialized and trained in an end-to-end form.
We use RMSProp optimizer with gradient descent to train the model.
Instead of padding the sequences to the same length,
we use the sequences with same number of visits to form a training batch.
The learning rate is set to 0.001.
Dimension of the medical code embedding is 64.
The dimension of the LSTM hidden layer is set to 128.
%{\color{red}Since the medical records on different EHRs datasets are recorded with different frequencies,
The unit of $\Delta^t$ is set to~\textit{year} on the MIMIC-III dataset and~\textit{day} on the PPMI dataset respectively.
		
To evaluate the performance of the proposed model, we compare it with
the following baseline models: %We set same hyper-parameters for every model and more details are described below.
			
		\begin{itemize}
			\item \textbf{LSTM}: We use basic LSTM as a simple baseline model.
Without considering the irregular temporal impact and inter-correlations of EHRs data,
we feed the mean-pooled feature $\bar{\bm{u}}^t$ and mean-concatenated feature $\bar{\bm{v}}^t$ into the LSTM instead of the attended feature $\hat{\bm{u}}^t$ and $\hat{\bm{v}}^t$.

			\item \textbf{LSTM+Att}: This model uses LSTM with attention mechanism which only considers the intra-sequence temporal unit without time-guided query.
				
			\item \textbf{T-LSTM} \cite{baytas2017patient}: T-LSTM uses a decaying function of time interval to adjust previous memory cell $\bm{c}_{t-1}$ which affects current output in LSTM.
				We set T-LSTM as a baseline model which considers the characteristic of varying time intervals in EHRs sequences.
			
			\item \textbf{RETAIN} \cite{choi2016retain}: RETAIN uses two RNNs to model visit-level and variable-level attention. It could detect influential past visits and clinical variables.
			
			\item \textbf{LSTM+TGA}: This model uses LSTM with interactive attention mechanism which only considers the intra-sequence irregular temporality item $\lambda_{u,q}^t$ and $\lambda_{v,q}^t$ in Eq.~\eqref{eq:final weight}.
				
			\item \textbf{LSTM+CoA}: This model uses LSTM with interactive attention mechanism which considers only the inter-sequence correlation item $\lambda_{u,v}^t$ in Eq.~\eqref{eq:final weight}.
			
			\item \textbf{TGCoA}: This model uses LSTM with attention mechanism which considers both the intra-sequence irregular temporality item $\lambda_{u,q}^t$ and  $\lambda_{v,q}^t$, and the inter-sequence correlation item $\lambda_{u,v}^t$  in Eq.~\eqref{eq:final weight}.
				
			\item \textbf{TGHOA}: This is the proposed time-guided high-order attention model which considers all attention items as shown in Eq. (\ref{eq:final weight}).
				
		\end{itemize}

	\subsection{Result Analysis}
The prediction results obtained by all baselines
are measured by three evaluation metrics including Accuracy, AUC-PR and AUC-ROC.
Table~\ref{tab:performance results} shows the
experimental results on both MIMIC-III and PPMI datasets.
As shown,
The proposed TGHOA outperforms all other models on both datasets.
		
For the mortality prediction task on the MIMIC-III dataset,
LSTM+TGA performs better than LSTM+Att.
It indicates intra-sequence irregular temporality unit
could better capture the irregular temporal impact than LSTM+Att which do not consider time intervals of sequential data.
We also get better performance than RETAIN
Besides, LSTM+TGA has better performance than T-LSTM.
It shows that considering irregular temporal impact with time-guided attention is more effective.
LSTM+CoA has higher scores compared to LSTM model.
It indicates that considering the inter-correlation between two types of EHRs data via attention mechanism is helpful.
The model LSTM+CoA that incorporates
the intra-sequence irregular temporality unit and the inter-source correlation unit further improves the performance.
Lastly, the proposed model TGHOA that considers the time-guided high-order correlations obtains the best performance.		
For the parkinson ranking task on the PPMI dataset,
the proposed TGHOA has similar performance improvements over baseline models.
%We can also observe that {\color{red} ``````}

%			\begin{figure}[t]
%				\centering
%				\includegraphics[width=7.5cm]{att.pdf}
%				\caption{The attention weights for medical features.}
%				\label{fig:att}
%             \end{figure}

\begin{figure}[t]
	\centering
	\includegraphics[width=8cm]{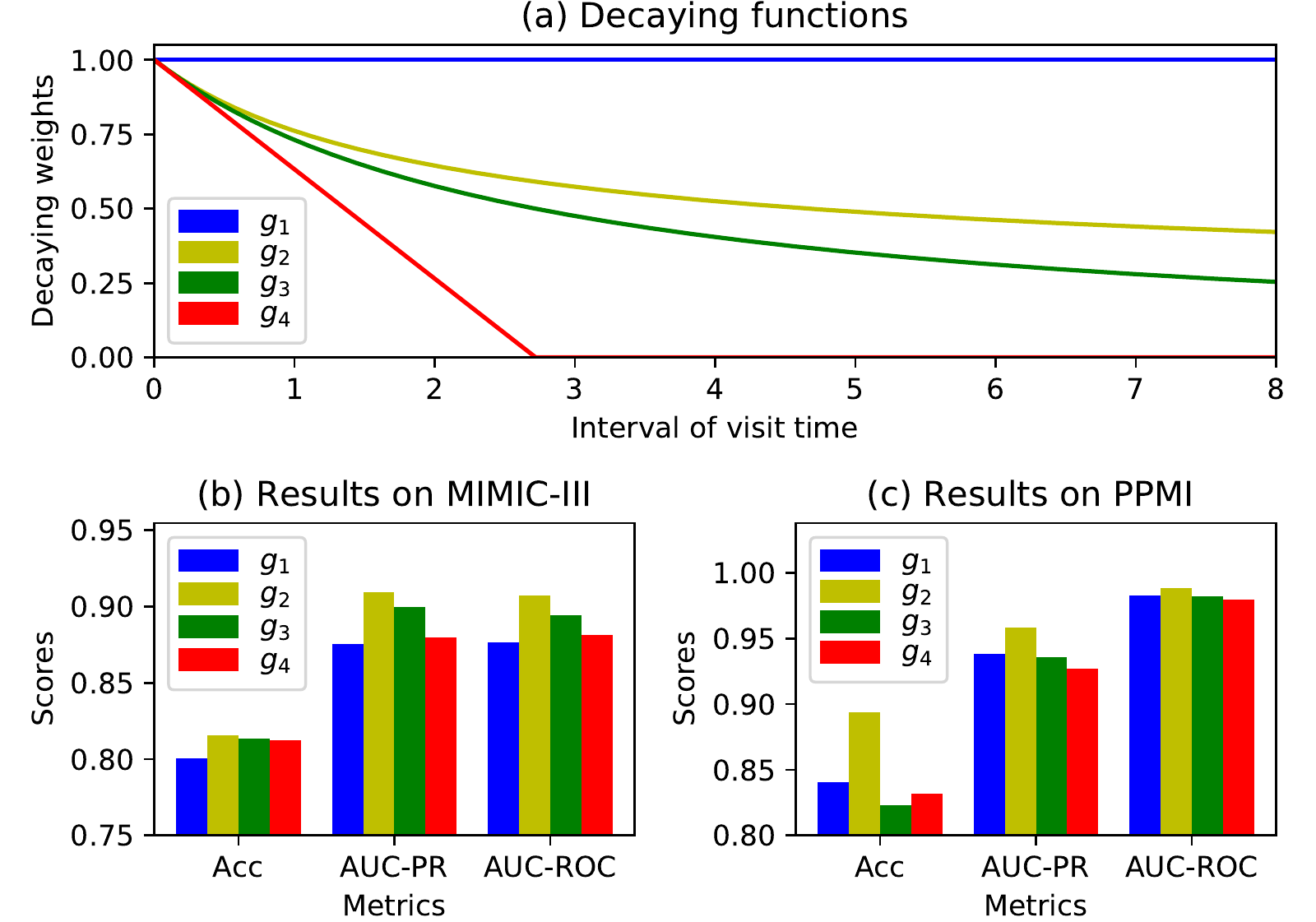}
	\caption{The effects of time-guided strategies.}
	\label{fig:time}	
	
\end{figure}
\subsection{Effects of Time-Guided Strategy}
In the proposed method,
the high-order attention module jointly considers the correlation between different modalities and
the irregular temporal impact of historical memory.
To further analysis the time-guided attention scheme,
we investigate the effects of different time-guided functions to the performance of TGHOA.
We compare four kinds of decaying functions
including $g_1(\Delta^t)=1$ without any decaying,
$g_2(\Delta^t)=1 / \log(\Delta^t + \mathrm{e})$,
$g_3(\Delta^t)=\mathrm{e} / (\Delta^t+\mathrm{e})$ and
$g_4(\Delta^t)=\max\{0, 1- \Delta^t / \mathrm{e}\}$.
Here, the $g_2$ is the adopted decaying function of the proposed method as introduced in Section~\ref{subsubsec:intra_seq}.
Figure~\ref{fig:time}(a) shows four function curves.
Note that the unit of $\Delta^t$ is ~\textit{year} on the MIMIC-III dataset and ~\textit{day} on the PPMI dataset respectively.
Figure~\ref{fig:time}(b) shows results obtained by our method with four decaying functions.
		
When using  $g_1$ as a guided function without time decaying,
our model obtains worst performance.
This further demonstrates that the time-guided attention scheme works well for modeling longitudinal EHRs data.
What's more, the decaying function $g_2$ performs better
than $g_3$ and $g_4$.
It indicates that if the attention model forgets the history feature too quickly,
we can only make a suboptimal health assessment, especially obvious on the PPMI dataset.

\begin{table}[t]
		\centering
		\caption{Diagnoses ranked according to attention scores.}
		\label{tab:case}
		\scalebox{0.8}{
		\begin{tabular}{cl}
			\toprule
			{Model}			&	Diagnoses (ICD-9 Code)			\\
			\midrule						
			TGHOA	&	Acidosis (276.2); History of  kidney neoplasm (V10.52); \\&Urinary complications (997.5);Atrial fibrillation (427.31); \\&Other noninfectious disorders of lymphatic channels (457.8);  \\&Other iatrogenic hypotension (458.29)\\
			\midrule						
			TGCoA	&	Other noninfectious disorders of lymphatic channels (457.8); \\&Acidosis (276.2);History of  kidney neoplasm (V10.52); \\&Atrial fibrillation (427.31); Urinary complications (997.5);    \\&Other iatrogenic hypotension (458.29)\\
			\midrule						
			LSTM+CoA	&Atrial fibrillation (427.31); Other iatrogenic hypotension (458.29); \\&Urinary complications (997.5); History of  kidney neoplasm (V10.52); \\&Acidosis (276.2); \\&Other noninfectious disorders of lymphatic channels (457.8); \\
			\midrule						
			LSTM+TGA	&	 Urinary complications (997.5); Other iatrogenic hypotension (458.29);\\& Atrial fibrillation (427.31); Acidosis (276.2); \\&History of  kidney neoplasm (V10.52); \\&Other noninfectious disorders of lymphatic channels (457.8);  \\
			\bottomrule
		\end{tabular}}
	\end{table}

\subsection{Case Study}

A key advantage of our model is its interpretability.
We conduct a case study of an unseen patient in the testing set of
the MIMIC-III dataset.
%Figure~\ref{fig:att} shows the attentions of medical features obtained by different baseline models.
%As shown,
%compared with other baselines,
%the proposed TGHOA focuses more on \textit{Acidosis (276.2)} and \textit{History of  kidney neoplasm (V10.52)} for diagnostic features, \textit{Heart Rate} and \textit{ABP} for physical features.
%This demonstrates that proposed attention mechanism gives reasonable cues of the medical features for
%the accurate mortality prediction.
%This demonstrates that proposed attention mechanism can identify the distinguished medical features which contribute more to the final prediction.

In Table~\ref{tab:case},
we rank the diagnostic codes according to their attention scores.
We could see that \textit{Acidosis (276.2)} and \textit{History of kidney neoplasm (V10.52)}, which have high fatality rate, have got high attention scores in TGHOA. While other diseases are complications which would not directly cause death. This results demonstrate that proposed attention mechanism gives reasonable cues of the medical features for the mortality prediction.
On the other hand, TGHOA and TGCoA generate very different attention scores of the medical feature \textit{Other noninfectious disorders of lymphatic channels (457.8)} while other diagnoses have similar rank. The LSTM+CoA and LSTM+TGA have distinctly different diagnostic attention ranks.
It shows that neither LSTM+CoA nor LSTM+TGA has modeled the complete correlation information of the EHRs data.

%		To demonstrate the effectiveness ofthe proposed attention mechanism in TGHOA, we plot the real-time attention weights of diagnostic codes of two patients. As show in Fig \ref{fig:a} for a surviving patient, it shows that a patient could survive with high probability although this patient has suffered high attention diseases in history, as long as long these symptoms were cured (e.g., code 29 and code 30), As show in Fig \ref{fig:b}, the patient with high risk of death suffers some diseases that are payed longitudinal high attention (e.g, code 4 and code 70).

\section{Conclusions}
In this paper,
we proposed a time-guided high-order attention (TGHOA)  model for analyzing the
heterogeneous and irregular temporal longitudinal EHRs data.
The diagnose features, physical indicator features and historical event features
were comprehensively used to
compute a relationship matrix
which was further transformed to
attention scores.
The irregular time interval was
used as an important factor to guide the attention computing.
The proposed high-order attention model
was evaluated on the MIMIC-III and PPMI datasets.
Extensive experimental results
demonstrated the effectiveness and interpretability of the proposed method.

~\\
\noindent\textbf{Acknowledgments.}
This work was supported in part by National Key Research and Development Program of China (No. 2017YFB1002804), National Natural Science Foundation of China (No. 61702511, 61720106006, 1711530243, 61620106003, 61432019, 61632007, U1705262, U1836220) and Key Research Program of Frontier Sciences, CAS, Grant NO. QYZDJSSWJSC039. This work was also supported by Research Program of National Laboratory of Pattern Recognition (No. Z-2018007) and CCF-Tencent Open Fund.

\bibliographystyle{splncs04}
\bibliography{pricai}
%Author, F.: Article title. Journal \textbf{2}(5), 99--110 (2016)
%Author, F., Author, S.: Title of a proceedings paper. In: Editor,
%F., Editor, S. (eds.) CONFERENCE 2016, LNCS, vol. 9999, pp. 1--13.
%Springer, Heidelberg (2016). \doi{10.10007/1234567890}
%Author, F., Author, S., Author, T.: Book title. 2nd edn. Publisher,
%Author, A.-B.: Contribution title. In: 9th International Proceedings
%on Proceedings, pp. 1--2. Publisher, Location (2010)
%LNCS Homepage, \url{http://www.springer.com/lncs}. Last accessed 4
%\end{thebibliography}
\end{document}